\documentclass{article}

\pdfoutput=1

\usepackage{arxiv}
\usepackage[utf8]{inputenc} % allow utf-8 input
\usepackage[T1]{fontenc}    % use 8-bit T1 fonts
\usepackage{hyperref}       % hyperlinks
\usepackage{url}            % simple URL typesetting
\usepackage{booktabs}       % professional-quality tables
\usepackage{amsfonts}       % blackboard math symbols
\usepackage{nicefrac}       % compact symbols for 1/2, etc.
\usepackage{microtype}      % microtypography
\usepackage{lipsum}		% Can be removed after putting your text content
\usepackage{graphicx}
\usepackage{natbib}
\usepackage{doi}
\usepackage{multirow}
\usepackage{subcaption}
\usepackage{svg}
\usepackage{float} 
\usepackage{graphicx}
 \usepackage{amsmath}

\title{Comprehensive Manuscript Assessment with Text Summarization Using 69707 articles}

\author{ {Qichen Sun}\\
	Peking University\\
	\texttt{qichen.sun@stu.edu.pku.cn} \\
	\And
        {Yuxing Lu} \\
	Peking University\\
	\texttt{luyx@stu.edu.pku.cn} \\
        \And
        {Kun Xia} \\
	Peking University\\
	\texttt{fangfangtu@stu.edu.pku.cn} \\
        \And
        {Li Chen} \\
	Fudan University\\
	\texttt{chenli\_imib@fudan.edu.cn} \\
        \And
        {He Sun} \\
	Peking University\\
	\texttt{hesun@pku.edu.cn} \\
        \And
        {Jinzhuo Wang} \\
	Peking University\\
	\texttt{wangjinzhuo@edu.pku.cn} \\
}

\renewcommand{\shorttitle}{\textit{arXiv} Template}

\hypersetup{
pdftitle={A template for the arxiv style},
pdfsubject={q-bio.NC, q-bio.QM},
pdfauthor={David S.~Hippocampus, Elias D.~Striatum},
pdfkeywords={First keyword, Second keyword, More},
}

\begin{document}
\maketitle

\begin{abstract}
	Rapid and efficient assessment of the future impact of research articles is a significant concern for both authors and reviewers. The most common standard for measuring the impact of academic papers is the number of citations. In recent years, numerous efforts have been undertaken to predict citation counts within various citation windows. However, most of these studies focus solely on a specific academic field or require early citation counts for prediction, rendering them impractical for the early-stage evaluation of papers. In this work, we harness Scopus to curate a significantly comprehensive and large-scale dataset of information from 69707 scientific articles sourced from 99 journals spanning multiple disciplines. We propose a deep learning methodology for the impact-based classification tasks, which leverages semantic features extracted from the manuscripts and paper metadata. To summarize the semantic features, such as titles and abstracts, we employ a Transformer-based language model to encode semantic features and design a text fusion layer to capture shared information between titles and abstracts. We specifically focus on the following impact-based prediction tasks  using  information of scientific manuscripts in pre-publication stage: (1) The impact of journals in which the manuscripts will be published. (2) The future impact of manuscripts themselves. Extensive experiments on our datasets demonstrate the superiority of our proposed model for impact-based prediction tasks. We also demonstrate potentials in generating manuscript's feedback and improvement suggestions.
\end{abstract}

% keywords can be removed

\section{INTRODUCTION}
In the 21st century, the substantial growth of research institutions has led to an almost exponential increase in the number of academic articles. As the volume of scientific articles surges, evaluating their impact swiftly and efficiently has become both intriguing and crucial. Predicting the future impact of these articles can guide reviewers in their assessments and help authors refine the research direction of their work. Moreover, assessing the potential future impact of multiple articles can offer insights into an author's prospective academic contributions, assisting academic institutions in their endeavors. Furthermore, many academic institutions now favor publications in esteemed journals. Thus, forecasting the potential influence of journals where articles might be accepted, based on their content, is a priority for both authors and reviewers.

Since the citation counts of articles can be easily obtained from databases, it has become one of the most commonly used metrics to measure the impact of articles. Numerous studies have attempted to predict the future citation counts of articles using various methods as an estimation of their future impact \cite{ma2021deep, tang2023predictable, vergoulis2020simplifying, li2019deep}. Currently, these studies can  be mainly divided into two categories: one uses regression models to predict the specific citation counts of articles \cite{ma2021deep, li2019deep}, while the other uses classification models to simply identify highly cited articles in the future \cite{tang2023predictable, vergoulis2020simplifying}. Scholars have utilized various features to accomplish these predictions, including semantic features of articles, bibliometric features, author features, and early citation count features \cite{mahalakshmi2020neural}. Many researchers have analyzed the relationship between various features and the future citation counts of articles to construct more effective prediction models \cite{hassan2020predicting, ruan2020predicting, mammola2021impact}. But we noticed that many works only focus on the academic researches from a specific disciplinary field, which narrows their application and ties the citation prediction task to a relatively restricted area. 

In most previous researches on citation prediction, numerical data types, such as early citation counts of articles, publication date, the number of references in the article, and some author-related data, have been explored at a high frequency, among which the early citation count has peaked as the prevalent numerical feature indicators. However, for those newly published papers, the temporary deficit of citation count makes it hard to support the prediction task, thus blocking the development of early stage article assessment through the early citation counts. In contrast, textual features of the article, such as the titles and abstracts, bear a richness in information and accessibility in obtaining. As the development of text mining greatly diminishes the difficulty of extracting the textual features from the article, it is important to explore how to more effectively utilize the textual features of articles for predicting their future impact.

Since previous studies have shown that there is a weak positive correlation between the Impact Factor (IF) of a journal and the exact citation counts of individual articles \cite{bozzo2017journal, asaad2020citation}, existing works on predicting future citation counts of articles can hardly accomplish the task of impact prediction for journals in which articles can be published. Although it is valuable to predict whether an article can be accepted by high-impact journals in addition to predicting the citation count of an article, there are currently no formal works that predict the impact of the corresponding journal  based on the information within the article itself to the best of our knowledge. 

In summary, the current challenges in predicting the impact of articles are as follows:
\begin{itemize}
\item  Many works focus on specific fields to achieve better prediction results. Can we develop a more general prediction model by using article data in multidisciplinary journals?
\item  There are few works about predicting the impact of the journal where the article is published. Can we provide reasonable standards for evaluating journals and the impact of articles and then complete the prediction task with aforementioned standards?
\item   How can we complete the prediction task only using features that can be obtained at the pre-publication stage? How can we extract important information more effectively from the text of articles and apply it to the prediction task?
\end{itemize}

In this article, we construct a large-scale database from multidisciplinary journals and propose the metrics for the impact-based tasks for journals in which articles will be published and manuscripts. We propose a model architecture based on  the text features  and bibliometric features of articles for both tasks. The contributions of this article are as follows:

\begin{itemize}
\item We construct a dataset including 69,707 articles from 99 journals across multiple disciplines from 2015 to 2019. All the article features in this dataset can be obtained at the pre-publication stage, enabling the model to be applicable in the early stages of article review and helping to select high-quality papers for journals.

\item  We develop an impact-based classifier by leveraging information available exclusively in the pre-publication stage, employing various techniques for in-depth feature extraction and fusion. Our framework has demonstrated a high level of accuracy in performing impact-based classification tasks for both journals and articles.

\item In addition to predicting the impact of scientific papers, we also focus on the task of predicting the impact of the journal where the article is published, which, to the best of our knowledge, has received little attention in this field. For both tasks, we present reliable metrics to evaluate impact across multiple disciplines.
\end{itemize}

\section{RELATED WORKS}
Since the introduction of  citation counts as a pivotal indicator for article evaluation, numerous studies have emerged focusing on predicting the citation trends of newly-published articles. These endeavors primarily revolve around the exploration of article-based features and an array of computational methodologies.

A prominent factor frequently highlighted in predicting potential citation growth is the influence of early citations. Bornmann et al. \cite{bornmann2014improve} utilized a truncated citation window to encompass early citations and other variables to enhance citation impact predictions. Newman et al. \cite{newman2014prediction} later integrated newly collected citation data as an early citation variable, thereby refining the citation prediction task. To offer a more nuanced understanding of early citation-based features, Ying et al. \cite{chen2023article} pioneered the Article Scientific Prestige metric (ASP), employing eigenvector centrality and Jacobi iteration, processing data from over 63 million articles and a billion citations.

Additionally, the impact factor of the journal in which an article is published is often considered alongside early citations. For instance, Stegehuis et al. \cite{stegehuis2015predicting} devised a quantile regression model using two predictors: the journal's impact factor and early citations of the publication. Similarly, Abramo et al. \cite{abramo2019predicting} integrated the impact of the hosting journal and the received citations, utilizing linear regression models on a sizable dataset.

A subset of research suggests bibliographic features could hold the key to citation trends. Hassan et al. \cite{hassan2019influential} drew inspiration from Twitter user relationships, using a connected graph to model this relationship. The subsequent machine learning model they developed capitalizes on the influence scores of Twitter users, which they posited as a paramount feature for academic paper citation predictions. Meanwhile, Wang et al. \cite{wang2019can} identified four publication factors and utilized Spearman correlation and Logistic regression for the prediction task.

Expanding on the bibliographic features, Wu et al. \cite{wu2019predicting} extracted academic paper information from three fields, whereas Mahalakshmi et al. \cite{mahalakshmi2020neural} emphasized the article's inherent contribution to science, extracting both structural and semantic features for citation predictions. Li et al. \cite{li2019deep} ventured into deep learning techniques for mining bibliographic features, correlating the results using a Pearson coefficient.

Beyond citations and bibliographic features, some studies also incorporate metadata features to enhance prediction richness. For example, Sohrabi et al. \cite{sohrabi2017effect} introduced novel metrics such as ‘abstract ratio’ and ‘weight ratio’ before integrating them into a regression model. In contrast, Fronzetti et al. \cite{fronzetti2020predicting} leveraged text mining and social network analysis, achieving a commendable 80\% accuracy rate in citation predictions six years post-publication. Hu et al. \cite{hu2020identification} employed a latent Dirichlet allocation method, while Tang et al. \cite{tang2023predictable} sought a holistic description of academic information by amalgamating various features and machine learning models to identify potential high-impact papers.

\section{DATASET}
We present a dataset comprising paper information, which includes both bibliometric and text features, sourced from journals spanning multiple disciplines. While the metadata selected for research may appear straightforward to obtain, constructing a dataset is challenging. It necessitates high-quality data, consistent formatting, and the absence of missing values. To ensure our work is both comprehensive and credible, we opted to use Scopus \footnote{\href{https://www.scopus.com/}{https://www.scopus.com/}}, a platform trusted by over 3000 academic, government, and corporate institutions. Scopus stands as the world's largest scientific literature database, encompassing journals, books, and conference proceedings. Another compelling reason for selecting Scopus is its robust author name disambiguation algorithm, allowing us to accurately match author features to their respective papers.

\subsection{Feature selection}
We use the pybliometrics python library \cite{rose2019pybliometrics} to obtain the bibliometric features and text features, the details of selected features are shown in Table \ref{tab:1}. Our proposed feature framework encompasses four categories: text features, date features, reference features, and author features. Text features are derived from the titles and abstracts, as these typically encapsulate the primary contributions of papers. We categorize journals based on their JCR scores, treating the top 10\% as impactful journals and calculate the proportion of the reference papers published in impactful journals as ‘impact\_reference’ in Table \ref{tab:1}. In our dataset, the publication year serves as the date feature. For unpublished articles evaluated using our model, the estimated year of publication can be used. As evident in Table \ref{tab:1}, all chosen features can be readily acquired during the early stages of article completion, even prior to its publication.

\begin{table*}
\caption{Selected Features: We simultaneously consider the relevance of features to the prediction tasks and the difficulty of data acquisition, resulting in the identification of the following four types of features. All the features can be easily obtained in the pre-publication stage of articles, which allows our work to provide assistance in the early-stage review process of research papers.}
\label{tab:1}
\renewcommand{\arraystretch}{1.2} 
\setlength{\tabcolsep}{10pt}
\begin{tabular}{|c|c|l|}
 \hline
Feature Name      & Feature Type                        & Descriptions                                      \\ \hline
abstract          & \multirow{2}{*}{text features}      & The abstract of the article                                            \\ \cline{1-1} \cline{3-3} 
title             &                                     & The title of the article                                               \\ \hline
year              & date features                       & The year of publication of the article                                 \\ \hline
reference\_count  & \multirow{3}{*}{reference features} & Number of reference of the article                                     \\ \cline{1-1} \cline{3-3} 
reference\_age    &                                     & The average year of publication of references                          \\ \cline{1-1} \cline{3-3} 
impact\_reference &                                     & The proportion of the reference papers published in impactful journals \\ \hline
h\_index          & \multirow{3}{*}{author features}    & The h-index of the first author                                        \\ \cline{1-1} \cline{3-3} 
author\_cit       &                                     & The total citations of the first author                                 \\ \cline{1-1} \cline{3-3} 
author\_papers    &                                     & Number of papers of the first author                                   \\ \hline
\end{tabular}
\end{table*}
\subsection{Statistics of dataset}
\begin{figure}
	\centering

        \caption{Distribution of journals in our dataset.}
	\begin{subfigure}{0.47\linewidth}
		\centering
		\includegraphics[width=0.8\linewidth]{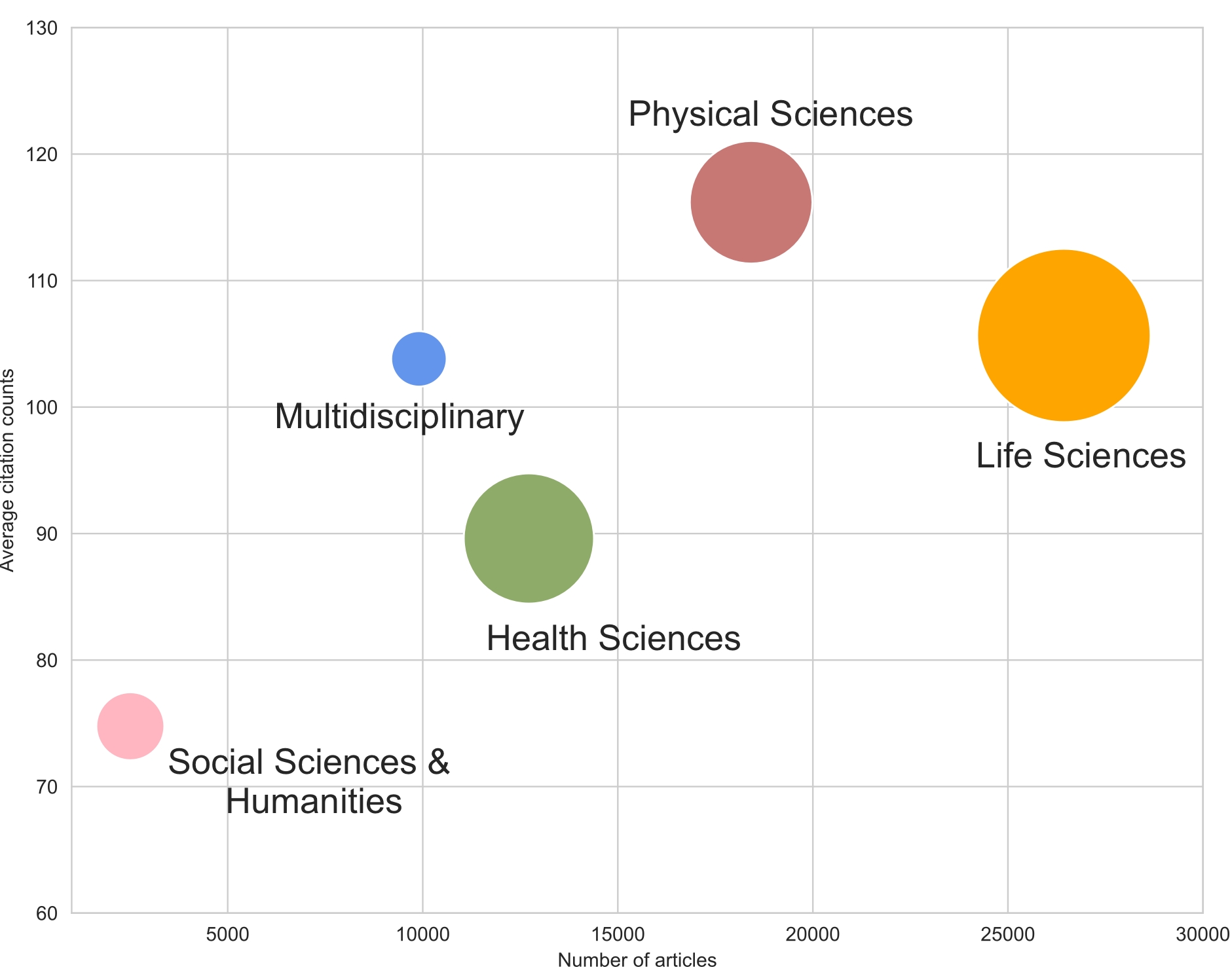}
		\caption{Distribution of the journals in five categories.}
		\label{distribution:1}
	\end{subfigure}
	\centering
	\begin{subfigure}{0.47\linewidth}
		\centering
		\includegraphics[width=0.78\linewidth]{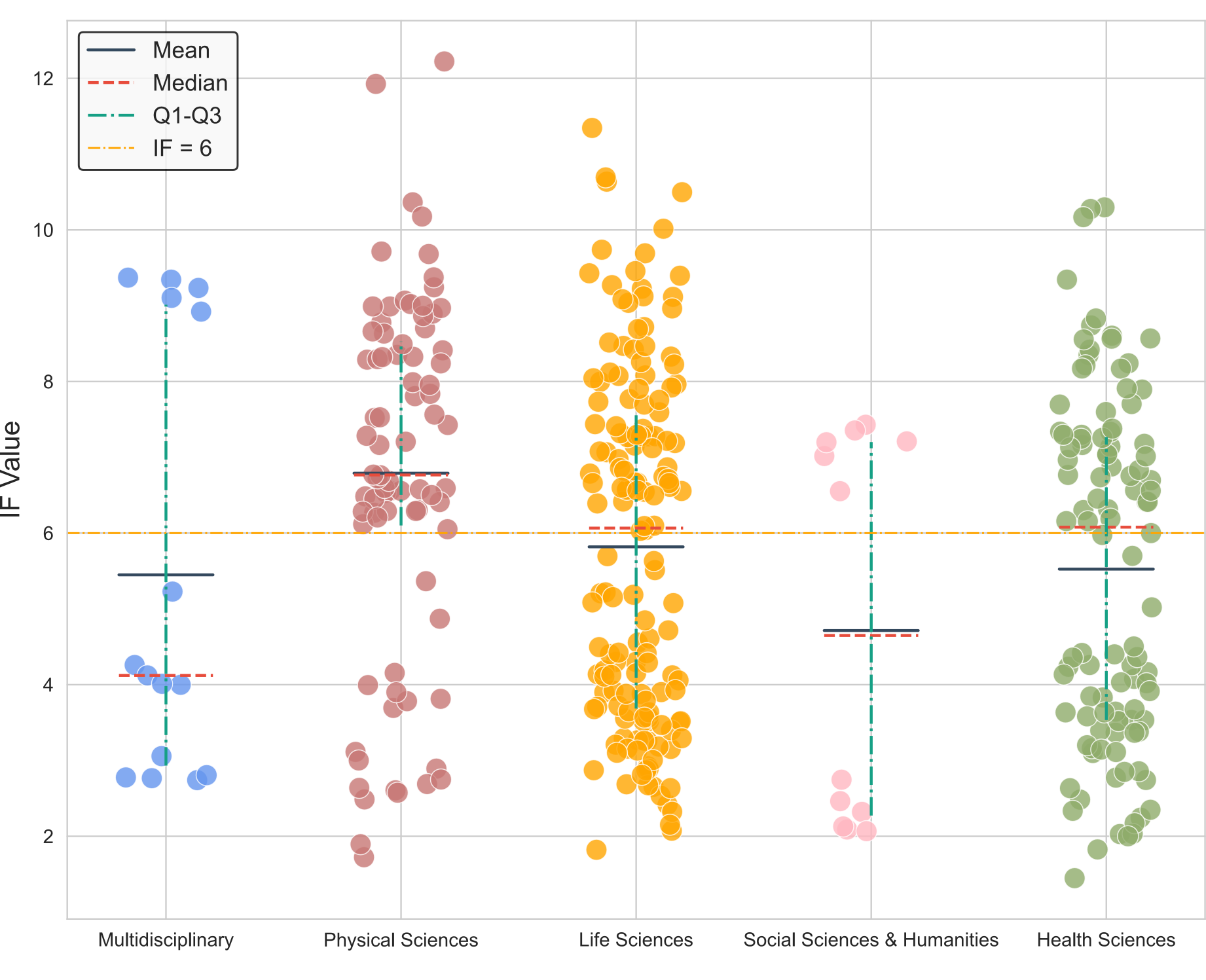}
		\caption{Distribution of IF scores of journals.}
		\label{distribution:2}
	\end{subfigure}
\end{figure}

The dataset consists of selected features of 69707 scientific articles which were published in 99 journals from 2015 to 2019 across multiple disciplines. We summarized the statistics of the dataset in Tabel \ref{tab:2} and the subject areas of journals in Fig \ref{distribution:1}. 
\begin{table*}[t]
\caption{Statistics of our dataset. Left two columns: Statistical summary of journals according to impact factor. Right two columns: Statistical summary of articles according to citations.}
\label{tab:2}
\renewcommand{\arraystretch}{1.2} 
\setlength{\tabcolsep}{21.8pt}
\begin{tabular}{|cc|cc|}
\hline
\multicolumn{2}{|c|}{Stastics of journals}                            & \multicolumn{2}{c|}{Stastics of articles}                        \\ \hline
\multicolumn{1}{|c|}{\#Journals}                              & 99    & \multicolumn{1}{c|}{\#Articles}                          & 69707 \\ \hline
\multicolumn{1}{|c|}{\#Articles in journals which IF $<$ 6}   & 39967 & \multicolumn{1}{c|}{\#Articles which citations $<$ 50}   & 45527 \\ \hline
\multicolumn{1}{|c|}{\#Articles in journals which IF $\ge$ 6} & 29740 & \multicolumn{1}{c|}{\#Articles which citations $\ge$ 50} & 24180 \\ \hline
\end{tabular}
\end{table*}

As we can observe in Table \ref{tab:2}, we select journals and articles  carefully to construct our dataset to ensure a balance in the quantity of  ‘high-impact’ journals (articles) and ‘others’ journals (articles). We focused on recent articles, spanning from 2015 to 2019, to capture the rapid advancements in the academic community. This approach contrasts with many existing studies that rely on older data.

Fig \ref{distribution:1} shows the data distribution of the five main disciplinary categories. We can conclude that:
\begin{itemize}
    \item Journals in the field of physical sciences tend to have the highest average citation counts. This is evidenced by the significant number of articles, as numerous well-acknowledged journals have been categorized under this field by Scopus.
    \item Journals in the field of life sciences  have the highest number of articles and  substantial average citation counts. This indicates a high research potential in this field in recent years, making it worthy of significant scholarly investment and efforts.
    \item  The multidisciplinary field has the fewest journals. This is primarily because most journal series only include a single journal in this category.
\end{itemize}

\section{METRICS AND PROBLEM FORMULATION}
Predicting the future impact of scientific articles holds significant importance for both authors and reviewers. In light of the recent surge in academic publications, there have been considerable attempts to accurately predict the future citation count for each article. However, in practice, we do not need to predict the exact number of future citations for papers in most situations. It is often sufficient to simply identify highly cited papers for most applications. Additionally, for many authors, the primary concern revolves around finding journals that would accept their manuscripts. Therefore, predicting the impact of the journals where articles might be published could be more meaningful to authors and reviewers during the early-stage evaluation of papers. In this paper, we focus on binary impact-based classification problems for both journals and scientific articles and propose reliable metrics for both tasks.

\subsection{Metrics for impact-based classification}
First, we would like to introduce the metric we use to identify high-impact journals --- Journal Impact Factor (JIF). JIF, which is commonly referred to as IF, gauges a journal's significance by calculating the number of times selected articles are cited within a specific year:
 \begin{equation}
  JIF=\frac{Citations_{T-1} + Citations_{T-2}}{Paper\_number_{T-1}+Paper\_number_{T-2}} 
\end{equation}
And we classify the journals as ‘high-impact’ journals if their $JIF \ge 6$ and the rest as ‘others’ journals.

\begin{figure}
    \centering
    \includegraphics[width=0.47\textwidth]{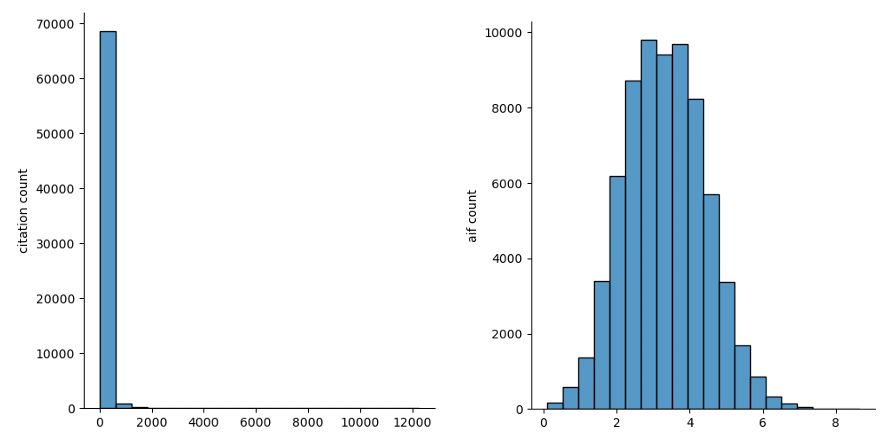}
    \caption{Distribution of the citation counts and the AIF.}
    \label{fig:dis}
    % \vspace{-10pt}
\end{figure}

As for identifying the high-impact articles, it is necessary to consider that the average citations of papers might be different across multiple disciplines so it may introduce bias by directly using citation count as a metric when evaluating the impact of articles in different fields. In order to ensure the universality of the metric for article impact, we employ IF to mitigate the influence arising from interdisciplinary variations. The final metric is calculated as follows:
\begin{equation}
    \begin{split}
        AIF&=\ln{(d \times cits\times p + (1-d) \times JIF)} \\
        p &= \min(2, \max(\frac{cits_m}{JIF}, 0.5))
    \end{split}
\end{equation}
where $cits$ is citation count of the article, and $cits_m$ is the median of the citation counts in our dataset, and $IF$ is Impact Factor of the corresponding journal, and $p$ is a scaling factor, and $d$ is a parameter to balance two terms of the equation.

The formulation of the aforementioned metric for article impact is based on three intuitions:
\begin{itemize}
\item Firstly, articles published in the same journal can have their impact evaluated based on citation counts. Consequently, within the same journal, the metric should increase in line with the article's citation count.
\item Second, the Impact Factors of journals in different fields can reflect the differences in citation counts for articles in these  corresponding fields. Therefore, we can apply IF to mitigate the disparities among articles from different disciplines. To realize this intuition, a scaling factor can be derived by dividing the median of the citation counts by the Impact Factor.
\item Thirdly, within a single field, there exists a spectrum of journals, ranging from ‘high-impact’ to ‘others’. To account for this variance, it's pivotal to factor in the nature of the journal where the article is published. One straightforward approach is to add the IF score to our metric.
\item Finally, as shown in Fig \ref{fig:dis}, citation counts often exhibit an extreme long-tail distribution. In our dataset, the maximum citation count for an article reached 12,249, while the median citation count was only 29. To prevent the AIF from also displaying an extreme long-tail distribution, we incorporated a logarithmic operation in our calculations.
\end{itemize}

Additionally, certain studies \cite{levitt2011combined} have integrated the IF with the article's own citation count to introduce a bibliometric indicator for evaluating the impact of articles. This further suggests that our metric for article impact classification is reasonable.

\begin{table}
\caption{Summary of results of the citation counts analysis. Only 622 of 34546 articles
underwent a change in classification during the 4 to  8 years after publication.}
\centering
\label{tab:cit}
\setlength{\tabcolsep}{20pt}
\begin{tabular}{ccc}
\toprule
Number of the year  & other articles & impactful articles\\ \midrule
4 years               & 30632  & 3914      \\
8 years               & 30010  & 4536      \\ \bottomrule
\end{tabular}
\end{table}

\subsection{The rationality of metrics}
Although we have introduced our intuitions of aforementioned metrics, there are still some specific details that we need to explain in conjunction with our dataset.

\subsubsection{Metric for journals}
 As for journals, we present a metric with fixed IF criterion, however, it may introduce bias when we assess the impact of journals across different scientific fields. To validate the rationality of this metric for journals, we plot a scatter plot of the IF of the journals in our dataset with the Impact Factor  as the y-axis and the categories of journal fields as the x-axis. From Fig \ref{distribution:2}, we can observe that there are both positive and negative examples of journals in each field when using $IF=6$ as the threshold.
 % 我把这里的metric改成了threshold，其他地方的metric可能有些也需要改一下
 Therefore, we can use the fixed IF criterion to assess the impact of journals across different scientific fields.

\subsubsection{Metric for articles}
For the metric for articles, using citation counts from the same time period for articles from different years may result in inaccurate classification. Can we use citation data in 2023 to evaluate impact of papers published from 2015 to 2019?  

To answer this question, we would like to explain that the error caused by using citation data in 2023 has a smaller impact on the AIF metric compared with using the citation counts only. Since, we choose $d \in (0, 0.5)$ as the balance parameter, we have the following inequality holds:
\begin{equation}
    dp \left |   cits_4 - cits_8\right |   < \left | cits_4-cits_8\right |
\end{equation}
where $cits_i$ is the citation counts in $i$ years after publication. Therefore, we can say that AIF is less sensitive about using citation data from the same time compared to citation counts. This allows us to concentrate on the distribution of citation counts 4 to 8 years post-publication, providing a rationale for using the citation data from 2023 for metric computation.

We analyze the distribution of citation counts in a public citation dataset called HEP-PH \cite{gehrke2003overview}. It covers all the citations within a dataset of 34,546 papers with 421,578 edges. We conduct a statistical analysis of the citation counts of articles in 4 to 8 years after publication. Articles receiving more than 20 citations are classified as ‘impactful’, while the others are classified as ‘others’. The results are presented in Table \ref{tab:cit} and we can observe that only 1.8\% of the articles underwent a change in classification during the 4 to  8 years after publication.  Therefore,  the small error for the binary classification caused by using citation data from the same time point is entirely acceptable, considering the cost of obtaining citation data at different time points. In summary, we could simply use citation data in 2023 to calculate the metric for the papers published from 2015 to 2019.

\subsubsection{Correlation analysis}
To verify the reliability of the selected impact assessment indicators, we conducted a correlation analysis using the bibliometric features available in the dataset, specifically the reference features and author features listed in Tab \ref{tab:1}. We believe that citation features reflect certain qualities of the papers themselves, while author characteristics allow us to assess the quality of their papers based on the author's past academic performance. Therefore, if the assessment indicators are reliable, they should have some correlation with these two types of features. We calculated the correlation coefficients between the features and the assessment indicators, and the results are shown in Tab \ref{tab:cor}.

\begin{table*}[ht]
    \centering
    \caption{Correlation analysis of various metrics with different features.}
    \label{tab:cor}
    \resizebox{0.85\textwidth}{!}{
        \begin{tabular}{@{}lccccccc@{}}
            \toprule
            & reference\_count & reference\_age & impact\_reference & h\_index & author\_cit & author\_papers \\ \midrule
            JIF          & 0.12 & 0.15 & \textbf{0.56} & 0.28 & 0.24 & 0.15 \\
            citations    & 0.20 & 0.10 & 0.25 & 0.26 & \textbf{0.27} & 0.14 \\
            AIF          & \textbf{0.29} & \textbf{0.18} & \textbf{0.47} & \textbf{0.31} & 0.16 & \textbf{0.16} \\ 
            \bottomrule
        \end{tabular}
    }
\end{table*}

From Tab \ref{tab:cor}, we can draw the following conclusions:
\begin{itemize}
    \item Considering the complexity of assessing paper quality, the feature of the proportion of references published in high-impact journals exhibits a striking correlation coefficient of 0.56 with the JIF. This may indicate that the high quality of references can, to some extent, reflect the level of work in the paper itself.
    \item The correlation coefficients for the citation count indicator are relatively low with almost all features, suggesting that directly using citation count as an assessment indicator is often unreliable.
    \item The AIF shows the highest overall correlation coefficients with other features, indirectly indicating the superiority of the assessment criteria we have chosen.
\end{itemize}

\subsection{Objective}
In summary, we can formulate the impact-based prediction problem as follows. Given the title, abstract and other metadata of a scientific article, our goal is to predict the future impact category of the article or the journal in which the article could be published. Each element in the test set $C_{test}$ consists the information of the scientific article $x_{i}$ and the corresponding ground-truth label $y_{i}$, i.e. $C_{test} = \left \{(x_{i},y_{i})  \right \} $ . The objective of the task is to learn a projection $f$ which can minimize the expected risk:
\begin{equation}
        f = \arg\min E_{(x,y)\sim C_{test}} \mathbf{1}(y\ne f(x)) 
\end{equation}
where $\mathbf{1}()$ is 0-1 loss function.

\section{METHOD}

%$\widetilde{\sigma}$ is the
%non-linearity map of the form: $\widetilde{\sigma}(v) = L_{1} \sigma( L_{2} %v)$, where $L_1$, $L_2$ are two linear layer, $\sigma$ is activation function.

\begin{figure*}[t]
    \centering
    \includegraphics[width=\textwidth]{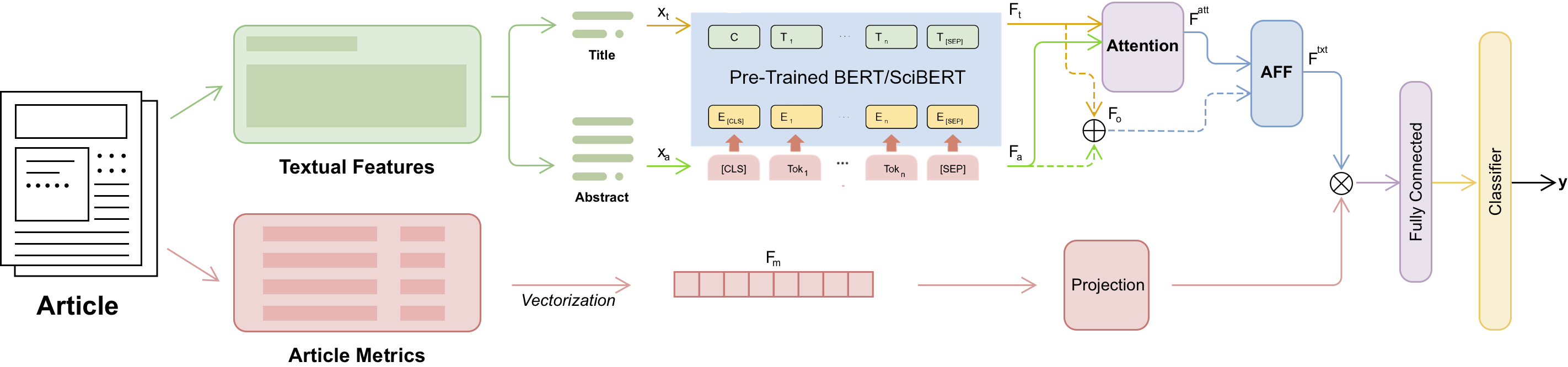}
    \caption{The overall architecture of IMAC model. It consists of text-encoding models, text feature fusion layers, metadata embedding layers and classifier.}
    \label{fig:pipeline}
\end{figure*}

The overall architecture of our Impact-based Manuscript Assessment Classifier (IMAC) model is shown in Fig \ref{fig:pipeline}. Given a tuple $(x_t, x_a, x_m)$ as input, where $x_t$ is title, $x_a$ is abstract and $x_m$ represents other metadata, we first use text-encoding model to encode the $x_t$ and $x_a$ respectively to extract text features $F_t$ and $F_a$. Then we fuse $F_t$ and $F_a$ by sending it to features fusion layers to obtain text feature $F^{txt}$. After that we send metadata to metadata embedding layers to extract $F^m$ and employ multiply operator to obtain a unified feature $F^u$. Finally, we send $F_u$ to the classifier to obtain the probabilities of our binary classification tasks. We could simply employ our model to the different classification tasks mentioned above by changing labels in the training set.
 
\subsection{Text encoder}
In our dataset $D$ , each element of $D$ is a tuple consisting of a title $x_t$ , an abstract $x_a$, and a metadata vector $x_m$. To extract features from $x_t$ and $x_a$, we need to represent them as token sequences using tokenizers.  After representation of $x_t$ and $x_a$, we could formulate the element send to our model as follows:
\begin{equation}
  \forall e_i \in  D,  e_i = (T_i, A_i, x_{m_i})
\end{equation}
where $T_i$ and $A_i$ represent the title and abstract sequence respectively, and $x_{m_i}$ is the metadata vector.
 
In our model, we used transformer-based pre-trained language model called SciBERT \cite{beltagy2019scibert} to deeply encode the token sequences of the $T_i$ and $A_i$ by mapping it to a $d$-dimension space, where $d=768$. SciBERT is a BERT model trained on a large corpus of scientific text and demonstrates significant improvements over BERT in many scientific NLP tasks. We choose SciBERT as our text encoder to extract features related to scientific domain. After SciBERT mapping text sequences onto a space of higher dimension, we employ a pooling layer to make the output of token embeddings into a single feature vector. Lastly, we apply a projected non-linearity of the form $\widetilde{\sigma}(v) = L_{1} \sigma( L_{2} v)$, where $L_1$ and $L_2$ are two linear layers acting space-map and $\sigma$ is activation function, in our case a GeLU function. Given the title and abstract sequence $T_i$ and $A_i$, the corresponding features are therefore formulated as follows:
\begin{equation}
    \begin{split}
        F_{t_i}&=\widetilde{\sigma}(SciBERT(T_i)) \\
        F_{a_i}&=\widetilde{\sigma}(SciBERT(A_i))
    \end{split}
\end{equation}
where $F_{t_i} \in \mathbb{R}^{768}$ and $F_{a_i} \in \mathbb{R}^{768}$.
\subsection{Text feature fusion}
As we have $F_{t_i}$ and $F_{a_i}$ after text encoding, we need to obtain a fused text feature for each article. To fuse title and abstract features of scientific articles thoroughly, we design a text fusion network to capture shared information between $F_{t_i}$ and $F_{a_i}$ in a high-level and filter out redundant information.

First, we employ an attention mechanism \cite{vaswani2017attention}, aiming at modeling the relationship and extracting shared information between title and abstract features. We could formulate the calculation process of the attention mechanism as follows:
\begin{equation}
    \begin{split}
        Att_{i}&=softmax\left ( \frac{F_{t_i}W_{Q}(F_{a_i}W_{K})}{\sqrt{d} }  \right ) F_{a_i}W_{V} \\
        F^{att}_{i}&=Drop(LayerNorm(Att_{i}))
    \end{split}
\end{equation}
where $Att_{i}$ is the feature extracted by attention mechanism with title features as queries, $W_{Q}, W_{K}, W_{V} \in \mathbb{R}^{768\times 768}$ are the weight matrices for the key, query, and value matrices respectively, $LayerNorm$ is the layer normalization for the attention mechanism, and $Drop$ is a dropout layer.

After obtaining text feature by employing attention mechanism, we simply obtain a $F^{aff}$  as follows inspired by ResNet \cite{he2016deep}, aiming to retain the important information from origin features:
\begin{equation}
    \begin{split}
        F_{o_i} &= F_{t_i} + F_{a_i} \\
        F^{aff}_i &= F_{o_i} + F^{att}_{i}
    \end{split}
\end{equation}

\begin{figure}[t]
    \centering
    \includegraphics[width=0.47\textwidth]{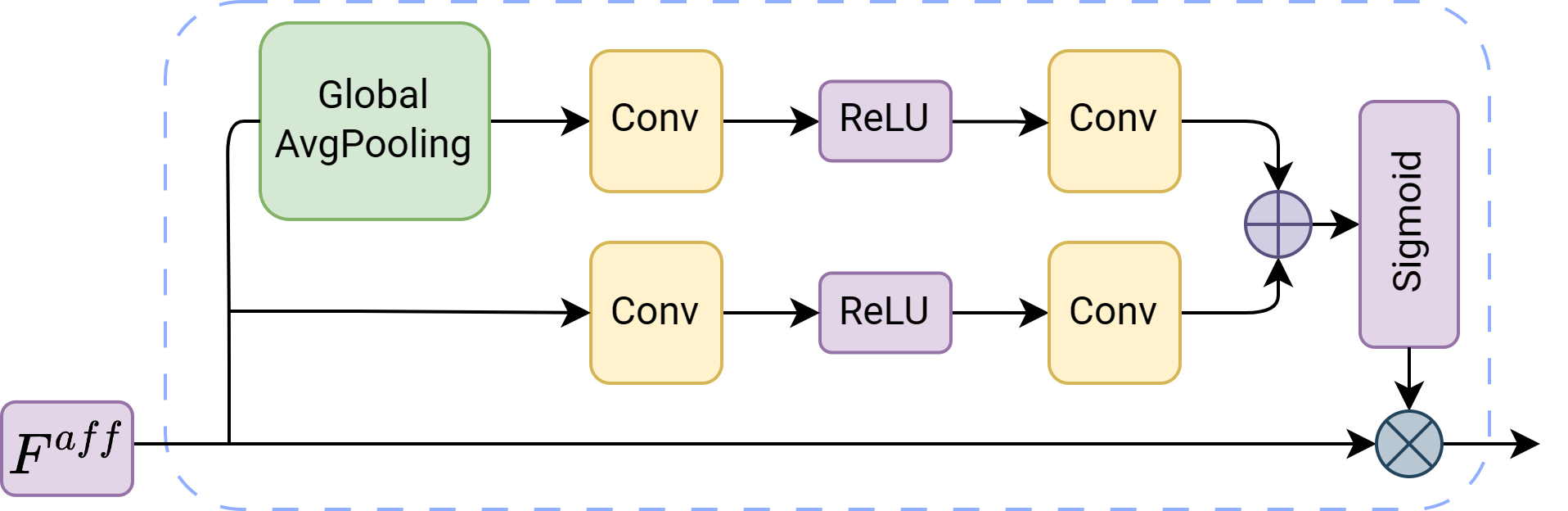}
    \caption{The architecture of MS-CAM mechanism in the AFF module.}
    \label{fig:MS}
    %\vspace{-5pt}
\end{figure}

Eventually, after the shortcut connection, we employ Attentional Feature Fusion (AFF) mechanism \cite{dai2021attentional} for higher-level feature fusion with $F_o$ as the identity mapping and $ F^{att}$ as the learned residual. The calculation process of AFF layer are formulated as follows:
\begin{equation}
    F^{txt}_{i} = M(F^{aff}_i) \otimes  F_{o_i} + (1 - M(F^{aff}_i)) \otimes F^{att}_{i}
\end{equation}
where $M()$ is the MS-CAM mechanism as shown  in  Fig \ref{fig:MS}, $\otimes$
denotes the element-wise multiplication, and $F^{aff}_i$, $F_{o_i}$, $F^{att}_{i}$ are reshaped to apply the MS-CAM mechanism, and $F^{txt}_{i} \in \mathbb{R}^{768}$ after reshaping it.

\subsection{Metadata fusion}
After obtaining the final text feature of each scientific article, we input metadata into our model to further improve the classification accuracy in this stage. 

First, we apply a fully connected neural network aiming at mapping metadata into a space of d dimension: 
\begin{equation}
    F^{m}_{i} = fc_{0}(x_{m_i})
\end{equation}
where $fc_{0}$ denotes fully connected neural network, and $F^{m}_{i}\in \mathbb{R}^{768}$.

Then we apply the element-wise multiplication between $F^{txt}$ and $F^m$ and employ another  fully connected neural network to obtain the final feature $F^u$:
\begin{equation}
        F^{u}_i = fc_{1}(F^{txt}_i \otimes F^{m}_i)
\end{equation}

\subsection{Class computation}
After obtaining the final feature $F^u$, we employ a linear classifier to allocate $F^u$ into the binary classes . Eventually, we can compute the prediction scores with softmax function as follows:
\begin{equation}
    \begin{split}
        out_i &= L(F^{u}_{i})  \\
        p_i =& softmax(out_i)
    \end{split}
\end{equation}

\subsection{Impact-based classifier learning}
After the discussion about our model architecture, we now discuss about the loss function we apply to train our prediction model in this section.

\subsubsection{Cross-entropy loss} 
As one of the most classic loss functions for classification tasks, cross-entropy loss allows our model's output to approximate the desired probability distribution effectively. We use cross-entropy loss as a component in our overall loss function, aiming to facilitate rapid learning of the approximate distribution during the early stages of training and enhance training efficiency. The calculation process of cross-entropy loss function is as follows:
\begin{equation}
    \mathcal{L} _c = - \sum _{i=1\dots N,j = \left \{  0,1\right \} }y_{i,j}\ln {p_{i,j}}
\end{equation}
where $y_{i,j}=1$ if the $i$-th ground-truth label is j, and $p_{i,j}$ is the prediction score of $i$-th element for $j$ class.

\subsubsection{Supervised contrastive loss}
To further learn from the extracted features, we assume that high-impact articles should have certain similarity in their extracted features. To learn from feature similarity, we employ the SupConLoss \cite{khosla2020supervised}. By computing the similarity between features of samples from the same category, we can further learn more detailed information related to article quality. Considering the characteristics of the binary classification task, we construct a mask matrix using the true labels of the samples to constrain the similarity of features with the same label. We can further enhance the generalization ability of our model by applying the constraint for similarity learning.

We use $I = \left \{ 1 \dots bs \right \} $ to denote a batch of elements we send to our model, $F^{u}_i$ is the final feature of the $i$-th element of the batch and $y_i$ for the class label. And then the SupConLoss function in our binary tasks can be formulated as follows:
\begin{equation}
    \mathcal{L} _s = -  \sum _{i \in I} \frac{1}{\left | P(i) \right | } \sum_{p\in P(i)} \log{\frac{\exp(F^{u}_{i}\cdot F^{u}_{p}/\tau)}{\sum_{a\in A(i)}\exp(F^{u}_{i}\cdot F^{u}_{a}/\tau)} }
\end{equation}
where $A(i) = I\backslash \left \{ i \right \} $, $P(i)= \left \{ p \in A(i): y_p = y_i \right \} $, $\tau$ is scaling factor.

\subsubsection{Overall loss function}
We combine the cross-entropy loss $\mathcal{L} _c$ and the SupConLoss $\mathcal{L} _s$ as the overall loss function:
\begin{equation}
    \mathcal{L} _o = \mathcal{L} _c + \alpha \mathcal{L} _s
\end{equation}
where $\alpha$ is a hyperparameter.

\section{EXPERIMENTAL RESULTS}
To demonstrate the effectiveness of our work, we show the results of experiments in this section.
\subsection{Settings}
The dataset presented in this paper utilizes data from the Scopus database, which is a subscription-based service provided by Elsevier. The use of the Scopus is authorized under the terms and conditions of our institutional subscription agreement with Elsevier.
\subsubsection{Copy rights}
Scopus is a registered trademark of Elsevier B.V. All rights in the Scopus dataset are reserved by Elsevier. The inclusion of data from Scopus in this research does not imply endorsement by Elsevier of the methods, findings, interpretations, or conclusions presented in this paper. We acknowledge and appreciate the valuable contributions of Elsevier and the authors of the original works included in the Scopus dataset. We have made efforts to appropriately cite and acknowledge the relevant sources within this paper.
\begin{table}[t]
\centering
\begin{minipage}{0.44\textwidth}

\caption{Experimental results for task 1:
identifying articles which can be published on the ‘high-impact’ journals.}
\label{tab:ex1}
\renewcommand{\arraystretch}{1.1} 
\begin{tabular}[width=0.8\textwidth]{ccccc}
 \toprule
Methods  & Accuracy & Precision & Recall & F1-score \\ \midrule
KNN      &   0.7031       &   0.7348        &  0.4686      &    0.5722      \\
SVM      &   0.6237       &    0.7524       &    0.1672    &    0.2736      \\
LR &    0.7543      &  0.7526         &  0.6261      & 0.6836         \\
ZeroR    &   0.5761     &  $-$         &  $-$      &   $-$       \\
\textbf{IMAC}      &      \textbf{0.9734}    &  \textbf{0.9752}   &  \textbf{0.9638}      &       \textbf{0.9684}   \\ \bottomrule
\end{tabular}
\end{minipage}
\hfill{
\begin{minipage}{0.44\textwidth}
\centering
\caption{Experimental results for task 2:
identifying ‘high-impact’ articles which have higher AIF than 'others'.}
\label{tab:ex2}
\renewcommand{\arraystretch}{1.1} 
\begin{tabular}[width=0.8\textwidth]{ccccc}
 \toprule
Methods  & Accuracy & Precision & Recall & F1-score \\ \midrule
KNN      &     0.7097     &    0.6879       & 0.4127       &  0.5159        \\
SVM      &    0.6626      &     0.7066      &    0.1701    &   0.2742       \\
LR &    0.7648      &   0.7070        &     0.6361   &  0.6697        \\
ZeroR    &      0.6253    &  $-$         &  $-$      &   $-$       \\
\textbf{IMAC}      &      \textbf{0.8438}    &  \textbf{0.7791}  &  \textbf{0.7985}      &       \textbf{0.7815}   \\ \bottomrule
\end{tabular}
\end{minipage}
}
\end{table}

\subsubsection{Dataset for experiments}
To conduct our experiments more effectively, we partitioned a small dataset to perform an ablation experiment in order to validate the effectiveness of different modules. And we conduct the comparative experiment with the complete dataset.

\subsubsection{Labeling and categorization}
For the task of predicting the impact of the journal where the article is published, we classify the journals as ‘high-impact’ journals if their $JIF \ge 6$ and the rest as ‘others’ . 
For the task of predicting articles' impacts, we classify articles as ‘high-impact’ articles if their $AIF \ge 5$ and the rest as ‘others’. 

\subsubsection{Assessment metrics}
The impact-based classification tasks are both binary classification tasks. Accuracy metric is commonly adopted for classification tasks and we also adopt Precision, Recall, and F1-score to evaluate the  performance more comprehensively.

\subsubsection{Model settings}
We test different pre-trained transformer-based models for extracting text features, i.e., BERT-base-uncased, Specter, and SciBERT. Eventually, we choose SciBERT for extracting features related to semantic information. Our model is implemented using Pytorch and trained with Adam optimizer. We use the initial learning rate of $1e-4$. For the $\mathcal{L} _o$, we choose $\alpha = 0.5$ to conduct the experiment and we choose $d=0.4$ for $AIF$. All reported results correspond to averages calculated over five runs.

\subsection{Baselines}
To investigate the effectiveness of the proposed model, we consider two types of baseline methods. The first type of baseline methods involves machine-learning based classifiers including KNN \cite{peterson2009knn}, SVM \cite{hearst1998supportsvm} and LR \cite{su2012linearlr}. They are trained with data objects represented by a prearranged set of features including the one-hot encodings of top-50-most-frequently used words in the titles and abstracts of scientific articles and the other article metadata. The second type is ZeroR, the simple feature-agnostic model, which always returns the most frequent category as prediction class.

\begin{figure}
    \centering
    \includegraphics[width=0.8\textwidth]{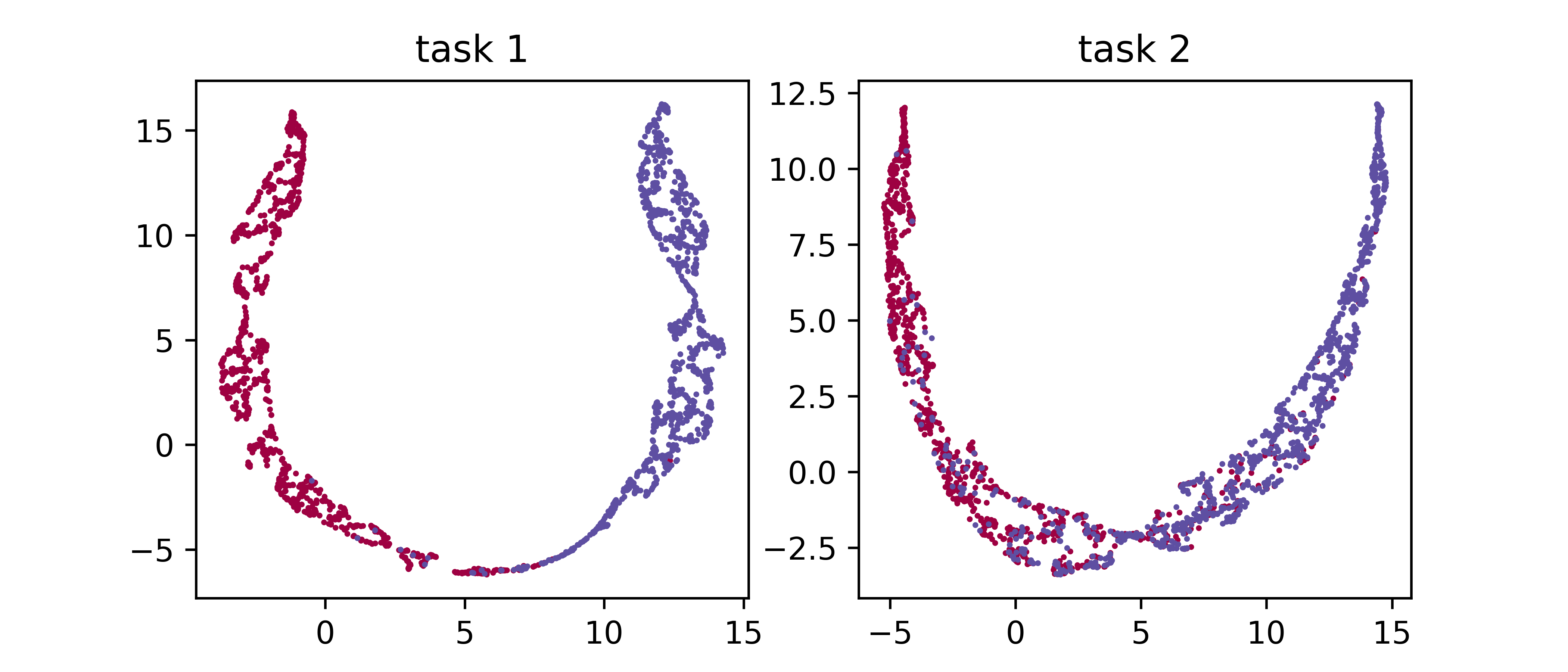}
    \caption{Visualization of the learned features on our dataset using UMAP. For each task, we randomly selected 1000 positive and negative examples.}
    \label{fig:tsne}
    % \vspace{-10pt}
\end{figure}

\subsection{Results}
Table \ref{tab:ex1} and Table \ref{tab:ex2} show the experiment results of our model and baseline methods for the two tasks. We have the following discussions by observing results:

\begin{itemize}
    \item For both tasks, all the baseline methods are consistently outperformed by our model (Accuracy of 0.9734 in task 1 and Accuracy of 0.8438 in task 2), which demonstrates the effectiveness of our model for impact-based classification tasks for scientific article assessment.
    \item ZeroR obtain the worst performance (Accuracy of 0.5761 in task 1 and Accuracy of 0.6253 in task 2) since it always returns the most frequent category without considering features. Therefore, it will obtain the unacceptable performance for most classification tasks.
    \item SVM exhibits the unacceptable performance with low recall (Recall of 0.1672 in task 1 and Recall of 0.1701 in task 2) for both tasks which shows that SVM tends to predict samples as negative instances.
    \item The LR outperforms all the competitors for both tasks (Accuracy of 0.7543 in task 1 and Accuracy of 0.7648 in task 2).
\end{itemize}

To further understand the effectiveness of our impact-based manuscript assessment classifier, we employ UMAP \cite{mcinnes2020umap} to visualize the learned features of our dataset. The results of the 2-D visualization for both tasks are shown in Fig \ref{fig:tsne}. For each task, we randomly selected 1000 positive and negative samples. As we can observe in Fig \ref{fig:tsne},  2 classes are clustered with clear boundaries for both tasks, which demonstrates the effectiveness of our proposed model.

\subsection{Ablation study}
Since obtaining relevant data for scientific research articles often requires the support of established databases which always do not allow
for large volume downloads, we are more concerned with the performance of the model on small-scale datasets in particular. As the proposed model contains multiple key components, we  conducted an ablation experiment to validate the effectiveness of our model on small-scale datasets. In the experiment, we compare variants of our model concerning the following aspects:

IMAC$\neg$ Sci: A variant of Our Model with Specter \cite{cohan2020specter}, another pre-trained BERT model with scientific text, instead of SciBERT. IMAC$\neg$ F: A variant of Our Model with text features fusing process being removed. IMAC$\neg$ LS: A variant of Our Model SupConLoss being removed.

\begin{table}[t]
    \caption{Ablation results on the IMAC model variants for tasks 1 and 2.}
    \label{tab:combined}
    \setlength{\tabcolsep}{7.5pt} % Adjust column spacing
    \begin{tabular}{c|cccc|ccccc} % Create a table with a vertical line separating the two tasks
        \toprule
        Models & \multicolumn{4}{c|}{Task 1} & \multicolumn{4}{c}{Task 2} \\ 
        \cmidrule(lr){2-5} \cmidrule(lr){6-9} % Draw horizontal lines under the headers
        & Accuracy & Precision & Recall & F1-score & Accuracy & Precision & Recall & F1-score \\ 
        \midrule
        IMAC$\neg$ Sci & 0.9282 & 0.9618 & 0.9335 & 0.9459 & 0.7507 & 0.8287 & 0.7648 & 0.7876 \\
        IMAC$\neg$ F & 0.9347 & 0.9506 & 0.9308 & 0.9405 & 0.7676 & 0.8434 & 0.7753 & 0.8020 \\
        IMAC$\neg$ LS & 0.9230 & 0.9483 & 0.9239 & 0.9356 & 0.7520 & \textbf{0.8938} & 0.7278 & 0.7991 \\
        \textbf{IMAC} & \textbf{0.9439} & \textbf{0.9654} & \textbf{0.9479} & \textbf{0.9557} & \textbf{0.7859} & 0.8628 & \textbf{0.7932} & \textbf{0.8199} \\
        \bottomrule
    \end{tabular}
\end{table}

As we can observe in Table \ref{tab:ab1} and Table \ref{tab:ab2}, the results of the ablation experiment show that:
\begin{itemize}
    \item IMAC outperforms IMAC$\neg$ Sci, indicating the effectiveness of SciBERT in semantic learning within scientific domains. By training on a substantial corpus of scientific text, SciBERT excels at extracting features relevant to scientific fields.
    \item IMAC outperforms IMAC$\neg$ F, which shows that the fusing process of text features contributes to prediction results. By applying attention mechanism and AFF mechanism, our model can capture the shared information between titles and abstracts and  filter out redundant information.
    \item IMAC outperforms IMAC$\neg$ LS, which indicates the effectiveness for similarity learning using supervised contrastive loss.
    The impact-based classifier optimizes $\mathcal{L}_s$ to learn from feature similarities within samples of the same category.
\end{itemize}

\begin{figure}[h]
    \centering
    \includegraphics[width=0.47\textwidth]{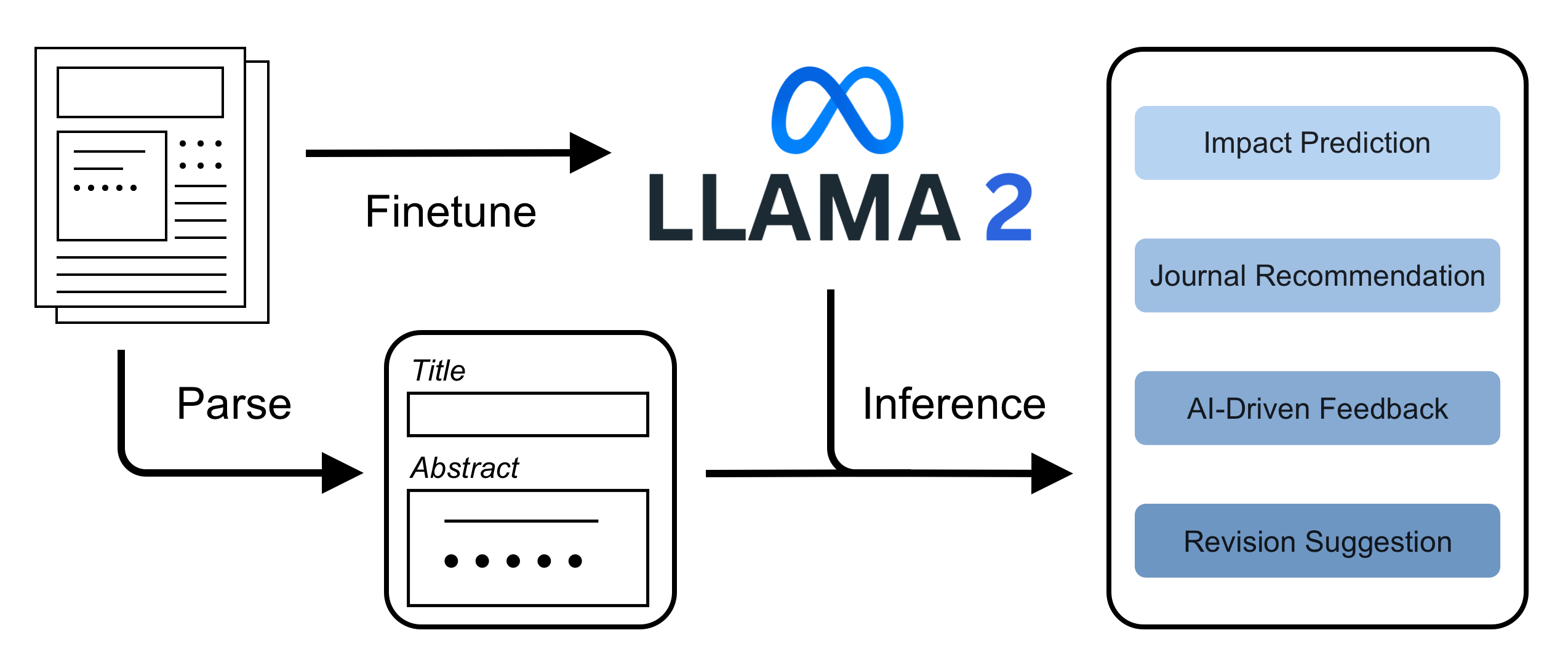}
    \caption{LLM-guided manuscript assessment and feedback.}
    \label{fig:fb}
    % \vspace{-10pt}
\end{figure}

\subsection{Discussions}
As mentioned above, we have demonstrated that our work can achieve satisfactory results in assessing the future impact of  manu\\-scripts. Furthermore, previous studies \cite{liang2023large} have validated the significant potential of large language models (e.g., GPT-4) in providing feedback and suggestions for articles. However, these studies only focused on a subset of 15 journals from the Nature series and data from ICLR in the years 2022-2023, which may limit the generalizability of their findings. In the future, we aim to further explore the potential of LLM in this field, aiming to provide more efficient feedback and suggestions for scholars as shown in Fig \ref{fig:fb}.

During the review stage of a manuscript, the feedback provided by reviewers is of paramount importance. Several platforms, such as OpenReview, have committed to crafting open-source platforms tailored for article critiques, facilitating authors in refining their contributions. We aim to leverage the powerful text summarization abilities of LLMs to help both authors and reviewers in obtaining a holistic assessment of the article's quality. In our future work, we will not only extract the article's title and abstract but also gather more comprehensive information about the article, such as the introduction, tables, and figures, to provide feedback on the overall paper. Additionally, it's worth noting that the ability of GPT-4 to generate specific and actionable feedback is limited. To address this limitation, we intend to train our own LLM using scientific texts to provide more specialized solutions. 

For authors, feasible suggestions for their articles can be more valuable. Both our work and Wang's work have already demonstrated the potential of deep learning techniques in this field. In the future, we will primarily focus on following two tasks:
\begin{itemize}
    \item Constructing a large-scale interdisciplinary database that includes a broader range of journals. This will enable us to provide suggestions on suitable journals for manuscript submission from a broader range.
    \item Paying attention to the content of the articles to provide assistance to authors in improving their research plans.
\end{itemize}
By expanding the database to offer guidance on manuscript submission and research improvement, we aim to provide authors with more valuable suggestions to enhance the quality of their articles. Additionally, it is important to note that expert human feedback is still irreplaceable despite the potential of LLMs in this field, as authors expect a fair assessment for their manuscripts \cite{redit2023curb}.

\section{CONCLUSION}
In this paper, we focus on the impact-based prediction tasks for the impact of journals in which manuscripts will be published and the impact of articles. First, we curate a significantly comprehensive and large-scale dataset sourced from Scopus, which includes information from 69707 articles of 99 multidisciplinary journals. Subsequently, we introduce metrics for both tasks and validate their reliability using our dataset. Eventually, we propose a novel Impact-based Manuscript Assessment Classifier (IMAC) model for both of the previously mentioned tasks. Comprehensive experiments conducted on our dataset highlight the superior performance of our proposed model in impact-based classification tasks.

\newpage

\bibliographystyle{unsrtnat}
\bibliography{references}  %%% Uncomment this line and comment out the ``thebibliography'' section below to use the external .bib file (using bibtex) .





% \end{thebibliography}

\end{document}